# Hyperspectral Image Denoising Employing a Spatial-Spectral Deep Residual Convolutional Neural Network

Qiangqiang Yuan, *Member, IEEE*, Qiang Zhang, *Student Member, IEEE*, Jie Li, *Member, IEEE*, Huanfeng Shen, *Senior Member, IEEE*, Liangpei Zhang, *Senior Member, IEEE*

*Abstract*—Hyperspectral image (HSI) denoising is a crucial preprocessing procedure to improve the performance of the subsequent HSI interpretation and applications. In this paper, a novel deep learning-based method for this task is proposed, by learning a non-linear end-to-end mapping between the noisy and clean HSIs with a combined spatial-spectral deep convolutional neural network (HSID-CNN). Both the spatial and spectral information are simultaneously assigned to the proposed network. In addition, multi-scale feature extraction and multi-level feature representation are respectively employed to capture both the multi-scale spatial-spectral feature and fuse different feature representations for the final restoration. The simulated and real-data experiments demonstrate that the proposed HSID-CNN outperforms many of the mainstream methods in both the quantitative evaluation indexes, visual effects, and HSI classification accuracy.

## I. Introduction

HYPERSPECTRAL images (HSIs), which simultaneously acquire both spatial and spectral information, have already been applied in many remote sensing applications, such as classification [1]–[2], target detection [3], unmixing [4], etc. Nevertheless, because of sensor internal malfunction, photon effects, and atmospheric interference, HSIs often suffer from various types of noise, such as random noise, stripe noise, and dead pixels [5]–[7]. This greatly affects the subsequent processing for information interpretation and understanding [8]–[10]. Therefore, it is critical to reduce the noise in HSIs and improve their quality before HSI analysis and interpretation.

A variety of HSI denoising methods have been proposed over the last decades [11]–[25]. The most fundamental strategy is to apply a conventional 2-D image denoising method to the HSI band by band. For example, non-local self-similarity (NSS) based methods such as block-matching and 3-D filtering (BM3D) [26] and weighted nuclear norm minimization (WNNM) [27] or learning-based methods such as expected patch log likelihood (EPLL) [28] can also be directly employed for HSI denoising. However, these band-by-band denoising methods usually lead to larger spectral distortion [22], since the correlation of the spatial and spectral information between the different bands is not simultaneously taken into consideration [23]–[25].

Therefore, from the point of view of combined spatial-spectral constraints, many scholars have jointly utilized the spatial and spectral information to reduce HSI noise [23]. Although these spatial-spectral HSI denoising methods can achieve relatively better results, the good performance must precisely tune parameters for each HSI [24]. This generates the unintelligent and time-consuming for different HSI data. Besides, because the noise exists in both spatial and spectral domain with unequal strength, these methods are insufficient to satisfy this complex situation, and tend to produce the over-smooth or spectral distortion in more complex noise scenario [30]–[31]. Therefore, it is significant to build a fast, efficient and universal framework to adapt to the different HSI data with different situations.

Recently, the deep learning theory [32] solving the complex problem with an end-to-end fashion can provide a prominent strategy to solve the mentioned insufficient of existing methods. This type methods exploit feature representations learned exclusively from abundant data, instead of hand-crafting features that are mostly designed based on domain-specific knowledge [33]. DL has also been introduced into the geoscience and remote sensing community for data interpretation, analysis, and application [34]–[35], including aerial scene classification [36]–[37], caption generation [38], synthetic aperture radar (SAR) image interpretation [39],

Manuscript received January 5, 2018; revised March 25, June 11, and July 7, 2018; accepted August 7, 2018. This work was supported in part by the National Key Research and Development Program of China under Grant 2016YFB0501403, in part by the National Natural Science Foundation of China under Grant 41701400, in part by the Fundamental Research Funds for the Central Universities under Grant 2042017kf0180, and in part by the Natural Science Foundation of Hubei Province under Grant ZRMS2016000241 and ZRMS2017000729. (Corresponding author: *Jie Li*.)

Q. Yuan is with the School of Geodesy and Geomatics and the Collaborative Innovation Center of Geospatial Technology, Wuhan University, Wuhan 430079, China (e-mail: qqyuan@sgg.whu.edu.cn).

Q. Zhang and J. Li are with the School of Geodesy and Geomatics, Wuhan University, Wuhan 430079, China (e-mail: whuqzhang@gmail.com; aaronleecool@whu.edu.cn).

H. Shen is with the School of Resource and Environmental Science and the Collaborative Innovation Center of Geospatial Technology, Wuhan University, Wuhan 430079, China (e-mail: shenhf@whu.edu.cn).

L. Zhang is with the State Key Laboratory of Information Engineering in Surveying, Mapping and Remote Sensing and the Collaborative Innovation Center of Geospatial Technology, Wuhan University, Wuhan 430079, China (e-mail: zlp62@whu.edu.cn).

pansharpening [40], and so on. In terms of nature image denoising task, some scholars such as Mao *et al.* [41] and Zhang *et al.* [42] employed convolutional neural networks (CNNs) to extract the intrinsic and different image features and avoid a complex priori constraint, which achieved state-of-the-art performance on nature image denoising. However, these denoising methods are lack of universality for HSI denoising, which do not consider the characteristics of spectral redundancy in HSI data. Therefore, how to combine with the spatial-spectral strategy and deep learning is significant for HSI denoising.

In this paper, considering that image noise in HSI data can be expressed through deep learning models between clean data and noisy data, we propose a combined spatial-spectral residual network with multi-scale feature extraction to recover noise-free HSIs. In our work, both the spatial structure and adjacent correlated spectra are simultaneously assigned to the proposed network for feature extraction and representation. The main ideas can be summarized as follows.

1) A novel spatial-spectral deep learning-based method for HSI denoising is proposed, by learning a non-linear end-to-end mapping between the noisy and clean HSIs with a 2D spatial and 3D spatial-spectral combined convolutional neural network. For better utilizing and mining the character of single band and high correction of its adjacent band, the proposed method develops a 2D and 3D combined convolutional neural network. In first layer of the proposed model, 2D-CNN can enhance the feature extraction ability of the single band, and 3D-CNN can simultaneously utilize the high correction and complementarity of its adjacent bands.

2) In remote sensing imagery, the feature expression may rely on contextual information in different scales, since ground objects usually have multiplicative sizes in different non-local regions. Therefore, the proposed model introduces a multi-scale convolutional unit to extract multi-scale features for the multi-context information, which can simultaneously get diverse receptive field sizes for noise removal.

3) For different HSIs with different spectrum numbers and diverse noise distributions, the proposed method can effectively remove the noise in different HSIs with only single model, which can simultaneously preserve the local details and structural information of the HSI without pre-set parameters adjusting.

The remainder of this paper is organized as follows. In Section II, the HSI degradation model is described, and then existing methods for HSI denoising is introduced. The proposed HSID-CNN model and the related details are presented in Section III. The simulated and real-data experimental results and a discussion are presented in Section IV. Finally, our conclusions are given in Section V.

## II. RELATED WORK

### A. Hyperspectral Noise Degradation Model

HSI data can be denoted by 3-D cube $\mathbf{Y}$ of size $M \times N \times B$, whose degradation model can be described as:

$$\mathbf{Y} = \mathbf{X} + \mathbf{V} \quad (1)$$

where $\mathbf{X}$ is the ideal noise-free data, $\mathbf{V} = [v_1, v_2, ..., v_B]$ is the additive noise with the Gaussian distribution $v_n \sim \xi(0, \sigma_n^2)$, and $1 \leq n \leq B$ and $\sigma_n^2$ mean that the noise intensity varies in the *n-th* spectra. Hence, the HSI denoising process is to estimate the original data $\mathbf{X}$ from the noisy observation $\mathbf{Y}$.

### B. Analysis of Existing HSI Denoising Methods

Up to now, there are two main types of HSI denoising methods: 1) transform-domain-based methods and 2) spatial-domain-based methods. The transform-domain-based methods attempt to separate clear signals from the noisy data by various transformations, such as principal component analysis, Fourier transform, or wavelet transform. For example, Atkinson *et al.* [11] presented an estimator utilizing discrete Fourier transform to decorrelate the signal in the spectral domain, and a wavelet transform was utilized for the spatial filtering. Othman *et al.* [12] employed a hybrid spatial-spectral derivative-domain wavelet shrinkage noise removal (HSSNR) method. This method depends on the spectral derivative domain, where the noise level is elevated, and benefits from the dissimilarity of the signal nature in the spatial and spectral dimensions. The major weakness of this type of approaches is that these methods are sensitive to the selection of the transform function and cannot consider the differences in the geometrical characteristics of HSIs.

To employ the reasonable assumption or prior, such as total variation [13], non-local [14]–[15], sparse representation [16]–[17], low rank models [18]–[22] and so on, the spatial-domain-based methods can map the noisy HSI to the clear one in attempt to preserve the spatial and spectral characteristics. For example, Yuan *et al.* [13] proposed a spatial-spectral adaptive total variation denoising algorithm. In addition, Chen *et al.* [14] also presented an extension of the (BM4D) [15] algorithm from video data to HSI cube data, with principal component analysis (PCA) for the noise reduction. Based on sparse representation, Lu *et al.* [16] proposed a spatial-spectral adaptive sparse representation (SSASR) method. Furthermore, Li *et al.* [17] exploited the intra-band structure and the inter-band correlation in the process of joint sparse representation and joint dictionary learning. For an HSI, both the high spectral correlation between adjacent bands and the high spatial similarity within one band reveal the low-rank structure of the HSI. Hence, Renard *et al.* [18] proposed a low-rank tensor approximation method, which performs both spatial low-rank approximation and spectral dimensionality reduction. In addition, Zhang *et al.* [19] proposed a new HSI restoration method based on low-rank matrix recovery (LRMR). Besides, Zhao *et al.* [20] investigated sparse coding to model the global redundancy and correlation (RAC) and the local RAC in the spectral domain, and then employed a low-rank constraint to consider the global RAC in the spectral domain. Instead of applying a traditional nuclear norm, Xie *et al.* [21] introduced a nonconvex low-rank regularizer named the weighted Schatten *p*-norm (WSN).

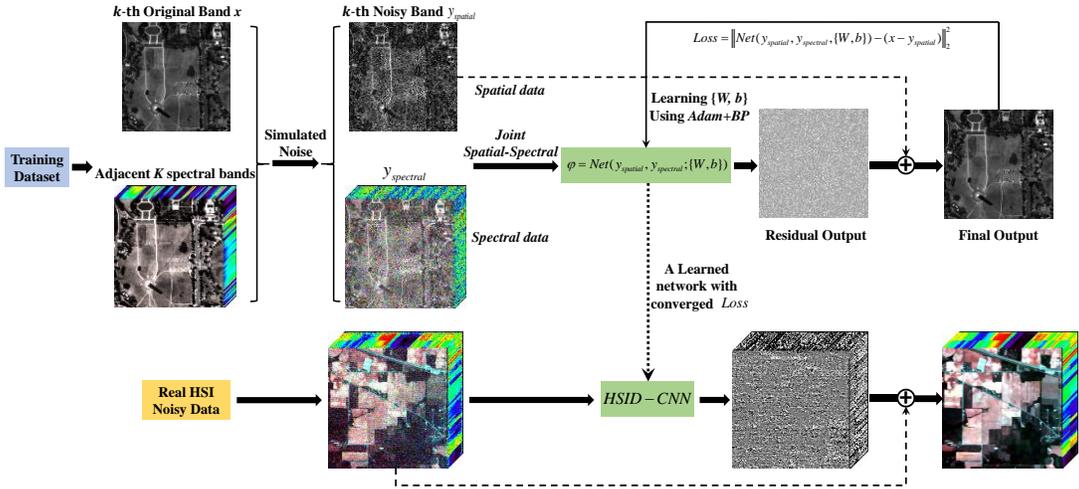

Fig. 1. Flowchart of the proposed HSID-CNN method for removing noise in HSI data.

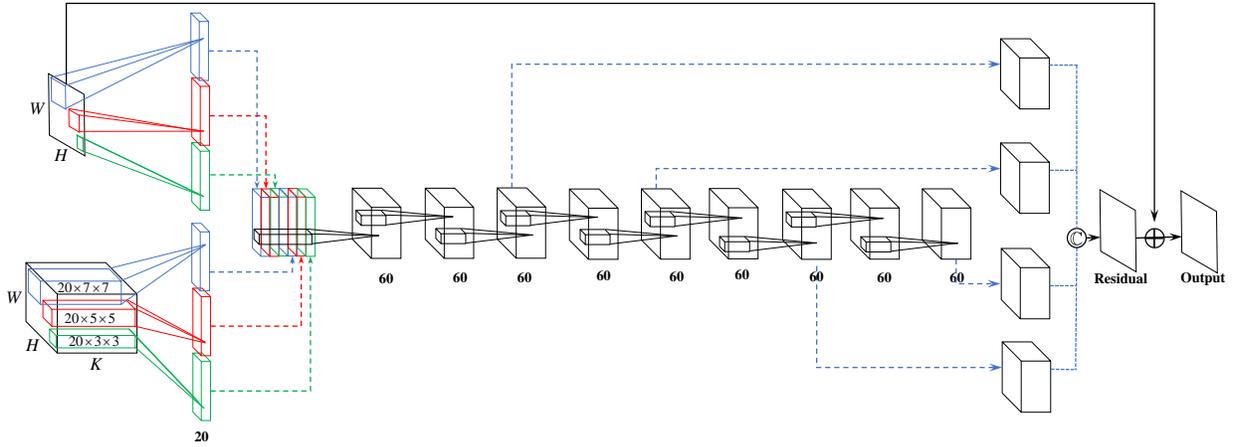

Fig. 2. Structure of HSID-CNN.

Although these HSI denoising methods can achieve relatively better results, the good performance must precisely tune parameters for each HSI [22]. This generates the unintelligent and time-consuming for different HSI data. Therefore, it is significant to build a fast, efficient and universal framework to adapt to the different HSI data with different situations.

## III. METHODOLOGY

### A. The Proposed Framework Description

Combined with the joint spatial-spectral strategy, we propose a novel DL-based method for HSI denoising with a deep CNN (HSID-CNN) to overcome the shortages of existing methods.

The flowchart of the proposed method is depicted in Fig. 1. HSID-CNN learns a non-linear end-to-end mapping between the noisy data and original data with a deep CNN, which simultaneously employs the simulated $k$-th noisy band $y_{spatial}$ and its adjacent bands $y_{spectral}$. The joint spatial-spectral data are then taken as the inputs of the proposed network, adaptively updating trainable parameters through the BP algorithm [33] with the residual output $\varphi$. After training with a converged loss, the learned network can be applied to the noise reduction for real HSI data. Details of this network are provided as below.

### B. The Proposed Model for HSI Denoising

The overall architecture of the HSID-CNN framework is displayed in Fig. 2. The input spatial data of size $W \times H$ represent the current noisy band in the top-left corner. Correspondingly, the input spectral data of size $W \times H \times K$ represent the current spatial-spectral cube with adjacent bands in the bottom-left corner. Based on this joint spatial-spectral learning strategy, one distinct advantage is that the proposed method can deal with no matter how many bands in HSIs data, because the proposed HSID-CNN model only takes one single spatial band (2D) as denoising object each time for HSI data, and its adjacent spectral bands (3D) as auxiliary data. Then our method traverses all the bands through one-by-one mode, which simultaneously employing spatial-spectral information with spatial and spatial-spectral filters, respectively. The detailed configuration of the proposed model is provided in Table I.

TABLE I
DETAILED CONFIGURATION OF HSID-CNN.

| Main parts | Configuration |
|---|---|
| Joint spatial-spectral multi-scale feature extraction | *Spatial_Feature_3*: $20\times3\times3$ Conv |
| | *Spatial_Feature_5*: $20\times5\times5$ Conv |
| | *Spatial_Feature_7*: $20\times7\times7$ Conv |
| | *Spectral_Feature_3*: $20\times3\times3$ Conv |
| | *Spectral_Feature_5*: $20\times5\times5$ Conv |
| | *Spectral_Feature_7*: $20\times7\times7$ Conv |
| | *Concat*: *Spatial feature + Spectral feature* |
| Feature representation | *Layer* 1–*Layer* 9: $60\times3\times3$ Conv + ReLU |
| Multi-level feature representation | *Concat*: *Layer* 3 + *Layer* 5 + *Layer* 7 + *Layer* 9 |
| Residual restoration | *Layer* 10: $1\times3\times3$ Conv |

**1) Joint Spatial-Spectral Multi-Scale Feature Extraction**

As mentioned in Section I, the redundant spectral information in HSIs can be of great benefit to improve the precision of the restoration, since the spatial-spectral cube usually has a high correlation and similarity in the surface properties and textural features. Therefore, for better utilizing and mining the character of single band and high correction with its adjacent band, the proposed method develops a 2D and 3D combined CNN network. In the proposed framework, the current spatial band and its $K$ adjacent bands are simultaneously set as the inputs in the proposed network. In Fig. 3(a) top, 2D convolution filters were employed to acquire spatial information for single current band. Simultaneously, in Fig. 3(a) bottom, 3D convolution filters (including adjacent spectrum numbers) were employed to acquire joint spatial-spectral information for adjacent bands.

Furthermore, the feature expression may rely on contextual information in different scales in remote sensing imagery, since ground objects usually have multiplicative sizes in different non-local regions. Therefore, the proposed model introduces a multi-scale convolutional unit to extract multi-scale features for the multi-context information, which can simultaneously get diverse receptive field sizes for noise removal. To capture both the multi-scale spatial feature and spectral feature, the proposed method employs different convolutional kernel sizes, as described in Fig. 3. The six outputs of the feature maps are then concatenated into a single 120-channel feature map. After extracting the contextual feature information with different scales, both the spatial information and spectral information can then be jointly utilized for *posteriori* processing.

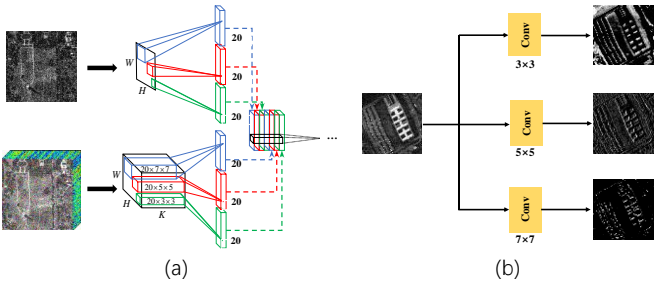

Fig. 3. Joint spatial-spectral multi-scale feature extraction. (a) The joint spatial-spectral multi-scale feature extraction block in proposed framework. (b) Multi-scale feature results with different convolution kernel sizes.

**2) Deep CNN with Residual Learning Strategy**

CNNs exploit the spatially local features by enforcing a local connectivity pattern between the convolutional junctions of adjacent layers. Hidden units in layer $l$ take as a subset of units in layer $l-1$, which form spatially contiguous receptive fields, obtaining more information by collecting and analyzing more neighboring pixels. Therefore, deeper networks can usually exploit the high non-linearity and obtain more essential feature extraction and expression abilities.

However, as the layer depth increases, the common deep networks can have difficulties in approximating identical mappings by stacked flat structures such as the *Conv-BN-ReLU* block [42]. In contrast, it is reasonable to consider that most pixel values in residual image for restoration will be very close to zero. In addition, the spatial distribution of the residual feature maps should be very sparse, which can transfer the gradient descent process to a much smoother hyper-surface of loss to the filtering parameters. Thus, it is significant to search for an allocation which is on the verge of the optimal for the network's parameters. Therefore, in the proposed model, the residual learning strategy is employed to ensure the stability and efficiency of the training procedure. The reconstructed output is represented with residual mode instead of straightforward results. Residual learning can effectively reduce the traditional degradation problem of the deeper networks [43], allowing us to add more trainable layers to the network and improve its performance. The residual noise $\varphi$ is defined as follows:

$$\varphi = \hat{x} - y_{spatial} \quad (2)$$

where $\hat{x}$ is the original clean band. Specifically, for the proposed HSID-CNN, given a collection of $T$ training image pairs $\{x^i, y^i_{spatial}, y^i_{spectral}\}_{i=1}^{T}$, $y^i_{spatial}$ is the noisy HSI as the spatial data, $y^i_{spectral}$ is the corresponding noisy adjacent cube as the spectral data, and $x^i$ is the clean HSI as the label. Setting $\Theta$ as the network trainable parameters, our model uses the mean-squared error (MSE) as the loss function:

$$loss(\Theta) = \frac{1}{2T} \sum_{i=1}^{T} \left\| Net(y^i_{spatial}, y^i_{spectral}, \Theta) - \varphi^i \right\|_2^2 \quad (3)$$

**3) Multi-Level Feature Representation for Restoration**

As shown in Fig. 4(b)–(e), various levels of feature information exist in the different depth layers. To efficiently utilize these comprehensive features between indirectly connected layers without direct attenuation, therefore, it is worth merging these different feature representations for the final restoration [44]. A multi-level feature representation unit in the proposed method is employed by integrally concatenating the multiple feature maps of the convolutional layers with different depths, as shown in Fig. 5.

Besides, multi-level representation in the proposed model can be regarded as multiple skip-connections [40], which have been verified the effectiveness for solving the vanishing gradient problem [42]. The concatenated representation is defined as:

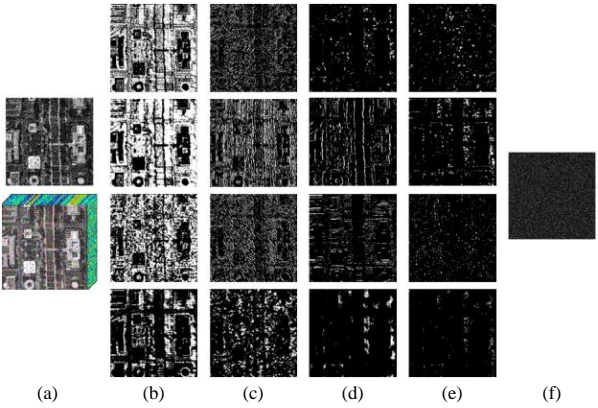

Fig. 4. Various levels of feature information in the different depth layers. (a) Input data with spatial/spectral images. (b) Feature maps of the 3rd convolutional layer. (c) 5th. (d) 7th. (e) 9th. (f) The output residual image.

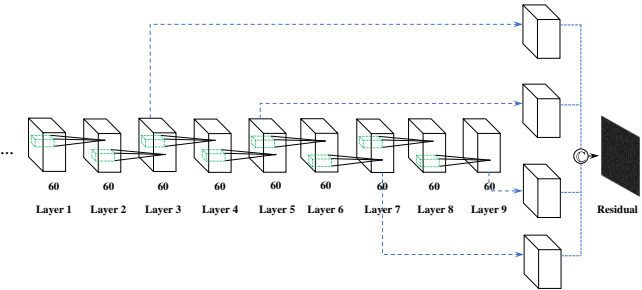

Fig. 5. Multi-level feature representation in the proposed HSID-CNN.

$$f_c = Concat\{f_3, f_5, f_7, f_9\} \quad (4)$$

where $f_3, f_5, f_7, f_9$ stand for the different-level feature representations, as displayed in Fig. 4(b)–(e), respectively. The concatenated layer $f_c$ is then further employed to fuse these combined feature representations for the final restoration:

$$\varphi = W_c * f_c + b_c \quad (5)$$

where $W_c$ and $b_c$ stand for the weight parameters and bias parameter of the last convolutional layer, respectively.

## IV. EXPERIMENTAL RESULTS AND DISCUSSION

To verify the effectiveness of the proposed method, both simulated and real-data experiments were performed, as described below. The proposed method was compared with the current mainstream methods of HSSNR [12], low-rank tensor approximation (LRTA) [18], BM4D [15], and LRMR [19]. Before the denoising process, the gray values of each HSI band were all normalized to $[0,1]$. MPSNR [45], MSSIM [46], and MSA [47] served as evaluation indexes in the simulated experiments. Generally speaking, in simulated experiments, better HSI denoising results are reflected by higher MPSNR, MSSIM, and lower MSA values. For the real-data experiments, the classification accuracy of the HSI before and after denoising is listed for comparison purposes with the different algorithms. The testing codes of the proposed method can be downloaded at https://github.com/WHUQZhang/HSID-CNN.

**1) Parameter Setting and Network Training:** The adjacent spectral band number $K$ was set as the same during all the training procedures, with $K = 24$ for both the simulated and real-data experiments. An impact analysis for the $K$ value is provided in Section IV-C. The proposed model was trained using the *Adam* [48] algorithm as the gradient descent optimization method, with momentum $\beta_1 = 0.9$, $\beta_2 = 0.999$, and $\varepsilon = 10^{-8}$, where the learning rate $\alpha$ was initialized to 0.01 for the whole network. The training process of HSID-CNN took 100 epochs (an epoch is equal to about 1,700 iterations, $batchsize = 128$). We employed the *Caffe* [49] framework to train the proposed HSID-CNN on a PC with 16 GB RAM, an Intel Xeon E5-2609 v3 CPU, and an NVIDIA Titan-X GPU. The training process for each model cost roughly 7 h 30 mins.

For training the proposed model, the Washington DC Mall image obtained by the Hyperspectral Digital Imagery Collection Experiment (HYDICE) airborne sensor [50], with the size of $1280 \times 303 \times 191$, was divided into two parts of $200 \times 200 \times 191$ for testing and other parts of $1080 \times 303 \times 191$ for training. These training data were then cropped in each patch size as $20 \times 20$, with the stride equal to 20. The simulated noisy patches are generated through imposing additive white Gaussian noise (AWGN) with different spectrums. The noise intensity is multiple and conforms to a fixed distribution or random probability distribution for different experiments. From the point of view of increasing the number of HSI training samples to better fit the HSI denoising mode, multi-angle image rotation (angles of 0, 90, 180, and 270 degrees) and multi-scale resizing (scales of 0.5, 1, 1.5, and 2 in our training data sets) were both utilized during the training procedure.

**2) Test data sets:** Three data sets were employed in the simulated and real-data experiments, as follows. The gray values of each HSI band were all normalized to $[0,1]$.

a) The first data set was the Washington DC Mall image mentioned above in Section IV-B, which was cropped to $200 \times 200$ for the simulated-data experiments. The image contained 191 bands after removing the water absorption bands.

b) The second data set was the AVIRIS Indian Pines hyperspectral image with the size of $145 \times 145 \times 220$, which was employed for the real-data experiments. A total of 206 bands was used in the experiments after removing bands 150–163, which are severely disturbed by the atmosphere and water.

c) The third data set was acquired by the ROSIS and covered the University of Pavia, Italy. The image scene is of $200 \times 200 \times 103$ after removing 12 water absorption bands.

### A. Simulated-Data Experiments

In the simulated HSI denoising process, the additional noise was simulated as the following three cases.

*Case 1*: For different bands, the noise intensity is equal. For example, $\sigma_n$ are set from 5 to 100, as listed in Table II.

*Case 2*: For different bands, the noise intensity is different and conforms to a random probability distribution (as shown in Table II '$\sigma_n = rand(25)$').

TABLE II
QUANTITATIVE EVALUATION OF THE DENOISING RESULTS OF THE SIMULATED EXPERIMENTS

| Noise level | Index | HSSNR | LRTA | BM4D | LRMR | Proposed |
|---|---|---|---|---|---|---|
| $\sigma_n = 5$ | MPSNR | 39.890 ± 0.0023 | 39.009 ± 0.0034 | 41.188 ± 0.0023 | 40.878 ± 0.0036 | **41.684 ± 0.0025** |
| | MSSIM | 0.9946 ± 0.0001 | 0.9926 ± 0.0002 | 0.9962 ± 0.0001 | 0.9952 ± 0.0001 | **0.9966 ± 0.0001** |
| | MSA | 2.3552 ± 0.0013 | 2.7008 ± 0.0015 | 1.9326 ± 0.0008 | 2.2760 ± 0.0011 | **1.8318 ± 0.0012** |
| $\sigma_n = 25$ | MPSNR | 28.018 ± 0.0024 | 30.672 ± 0.0033 | 31.136 ± 0.0025 | 33.029 ± 0.0023 | **33.050 ± 0.0028** |
| | MSSIM | 0.9361 ± 0.0001 | 0.9629 ± 0.0002 | 0.9685 ± 0.0002 | 0.9809 ± 0.0001 | **0.9813 ± 0.0001** |
| | MSA | 8.1332 ± 0.0034 | 5.7962 ± 0.0056 | 5.0514 ± 0.0048 | 4.6097 ± 0.0028 | **4.2641 ± 0.0026** |
| $\sigma_n = 50$ | MPSNR | 22.232 ± 0.0036 | 26.832 ± 0.0052 | 26.752 ± 0.0034 | 28.806 ± 0.0043 | **28.968 ± 0.0039** |
| | MSSIM | 0.8233 ± 0.0001 | 0.9246 ± 0.0001 | 0.9208 ± 0.0002 | 0.9532 ± 0.0001 | **0.9536 ± 0.0001** |
| | MSA | 14.413 ± 0.0048 | 7.4996 ± 0.0054 | 7.1405 ± 0.0056 | 6.8008 ± 0.0034 | **6.2197 ± 0.0045** |
| $\sigma_n = 75$ | MPSNR | 18.780 ± 0.0047 | 24.682 ± 0.0054 | 24.261 ± 0.0035 | 26.306 ± 0.0046 | **26.753 ± 0.0039** |
| | MSSIM | 0.7082 ± 0.0002 | 0.8866 ± 0.0001 | 0.8670 ± 0.0001 | 0.9192 ± 0.0001 | **0.9273 ± 0.0001** |
| | MSA | 19.904 ± 0.0053 | 8.4426 ± 0.0057 | 8.6010 ± 0.0064 | 8.5644 ± 0.0067 | **7.5246 ± 0.0052** |
| $\sigma_n = 100$ | MPSNR | 16.314 ± 0.0065 | 23.175 ± 0.0048 | 22.577 ± 0.0054 | 24.310 ± 0.0047 | **25.296 ± 0.0043** |
| | MSSIM | 0.6049 ± 0.0001 | 0.8494 ± 0.0003 | 0.8119 ± 0.0002 | 0.8799 ± 0.0002 | **0.9014 ± 0.0001** |
| | MSA | 24.732 ± 0.0065 | 9.1219 ± 0.0072 | 9.7611 ± 0.0068 | 10.468 ± 0.0074 | **8.4061 ± 0.0063** |
| $\sigma_n = rand(25)$ | MPSNR | 32.797 ± 0.0028 | 28.843 ± 0.0025 | 34.424 ± 0.0034 | 36.094 ± 0.0033 | **37.367 ± 0.0028** |
| | MSSIM | 0.9756 ± 0.0001 | 0.9331 ± 0.0001 | 0.9833 ± 0.0002 | 0.9856 ± 0.0001 | **0.9916 ± 0.0001** |
| | MSA | 5.0027 ± 0.0023 | 10.434 ± 0.0016 | 4.0766 ± 0.0027 | 3.4254 ± 0.0019 | **2.9571 ± 0.0026** |
| $\sigma_n = Gau(200,30)$ | MPSNR | 34.461 ± 0.0028 | 28.200 ± 0.0023 | 34.109 ± 0.0037 | 35.962 ± 0.0025 | **36.804 ± 0.0029** |
| | MSSIM | 0.9757 ± 0.0001 | 0.9119 ± 0.0002 | 0.9794 ± 0.0001 | 0.9893 ± 0.0001 | **0.9895 ± 0.0001** |
| | MSA | 5.1619 ± 0.0014 | 10.708 ± 0.0018 | 3.6714 ± 0.0012 | 3.4922 ± 0.0024 | **3.3156 ± 0.0017** |

*Case 3*: For different bands, the noise intensity is also different, where the noise level $\sigma_n$ is added along the spectral axis and is varied like a Gaussian curve [10].

$$\sigma_n = \beta \sqrt{\frac{\exp\{-(n-B/2)^2/2\eta^2\}}{\sum_{i=1}^{B} \exp\{-(i-B/2)^2/2\eta^2\}}} \quad (6)$$

where the intensity of the noise is restricted by $\beta$, with $\eta$ behaving like the standard deviation for the Gaussian curve. In the simulated experiments, the noise was defined as $\sigma_n = Gau(\beta, \eta)$, where $\beta = 200$ and $\eta = 30$.

To acquire an integrated comparison for the other methods and the proposed HSID-CNN, quantitative evaluation indexes (MPSNR, MSSIM, and MSA), a visual comparison, curves of the spectra, and the spectral difference results were used to analyze the results of the different methods. The averages and standard deviations of contrasting evaluation indexes of the three cases with various noise levels and types in 10 times are listed in Table II. To give detailed contrasting results, $\sigma_n = 100$, $\sigma_n = rand(25)$, and $\sigma_n = Gau(200,30)$ are chosen to demonstrate the visual results, corresponding to Fig. 6, Fig. 8, and Fig. 10, respectively. Due to the large number of bands in an HSI, only a few bands are selected to give the visual results in each case with pseudo-color. Fig. 6 shows the denoising results of the different methods in simulated Case 1 with the pseudo-color view of bands 17, 27, and 57 (see enlarged details in Fig. 7); Fig. 8 gives the denoising results of the different methods in simulated Case 2 (see enlarged details in Fig. 9); Fig. 10 shows the denoising results of the different methods in simulated Case 3 (see enlarged details in Fig. 11). The values of PSNR and SSIM within the different bands of the restored HSI are depicted to assess the per-band denoising result in Fig. 12. Furthermore, to verify the outputs from the spectral point of view, the spectral curves of the restoration results are displayed in Fig. 13. Meanwhile, the spectral difference curves of the roof, grass, and road classes are also given in Fig. 14, respectively.

In Table II, the best performance for each quality index is marked in bold and the second-best performance for each quality index is underlined. Compared with the other algorithms, the proposed HSID-CNN achieves the highest MPSNR and MSSIM values and the lowest MSA values in all the noise levels, in addition to showing a better visual quality in Figs. 6–12. Although the HSSNR algorithm has a good noise reduction ability under weak noise levels, as shown in Table II with $\sigma_n = 5$, it cannot well deal with strong noise levels such as $\sigma_n = 100$, and the results still contain obvious residual noise, especially in Figs. 6 and 7. LRTA performs well under the equal noise intensity for different spectra in Table II; however, it generates some fake artifacts in Figs. 6 and 7. From Table II, BM4D shows a good noise reduction ability, under both the uniform/non-uniform noise intensities for different bands. However, it also produces over-smoothing in the results in Figs. 6–11, since the different non-local similar cubes in the HSI may result in the removal of small texture features. By exploring the low-rank property of the HSI, LRMR also provides better denoising results. However, there are still some noise residuals in the magnified areas in Fig. 7, especially for the high noise intensity condition such as $\sigma_n = 100$ in Fig. 6.

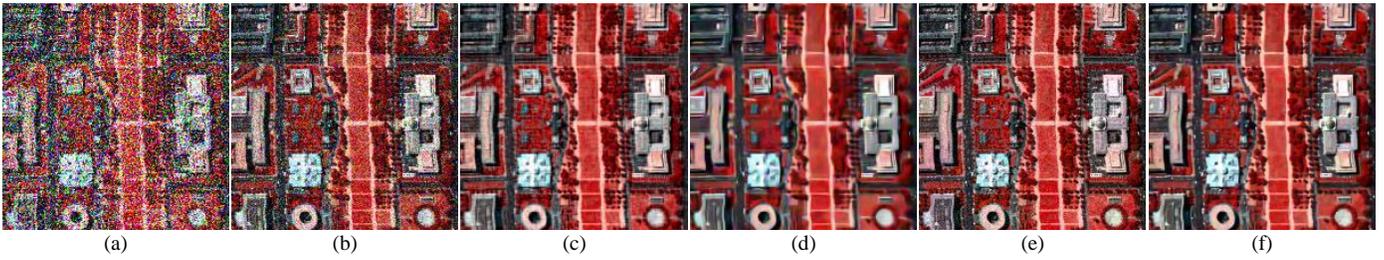

Fig. 6. Results for the Washington DC Mall image with $\sigma_n = 100$ in Case 1. (a) Pseudo-color noisy image with bands (57, 27, 17). (b) HSSNR. (c) LRTA. (d) BM4D. (e) LRMR. (f) The proposed method.

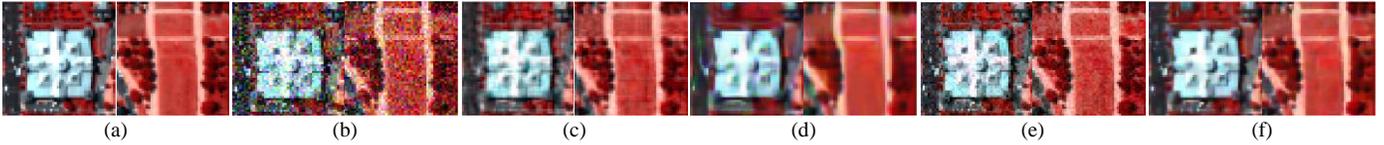

Fig. 7. Magnified results for the Washington DC Mall image in Case 1. (a) Noise-free image. (b) HSSNR. (c) LRTA. (d) BM4D. (e) LRMR. (f) The proposed method.

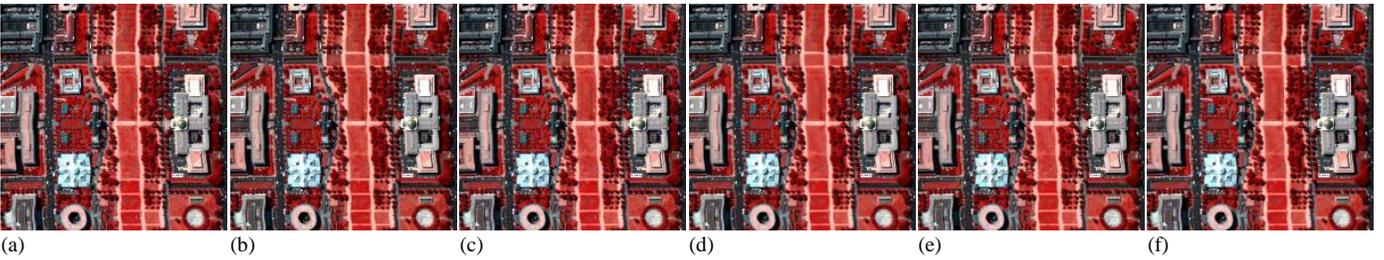

Fig. 8. Results for the Washington DC Mall image with $\sigma_n = rand(25)$ in Case 2. (a) Pseudo-color noisy image with bands (57, 27, 17). (b) HSSNR. (c) LRTA. (d) BM4D. (e) LRMR. (f) The proposed method.

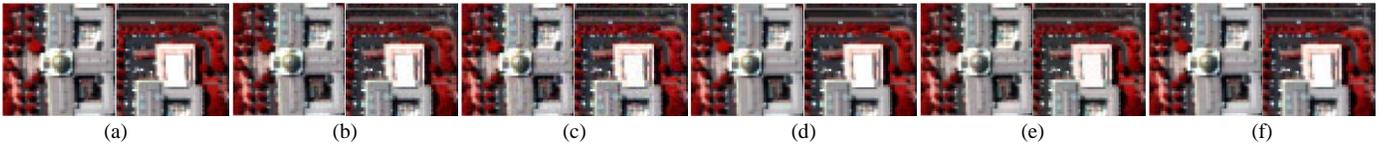

Fig. 9. Magnified results for the Washington DC Mall image in Case 2. (a) Noise-free image. (b) HSSNR. (c) LRTA. (d) BM4D. (e) LRMR. (f) The proposed method.

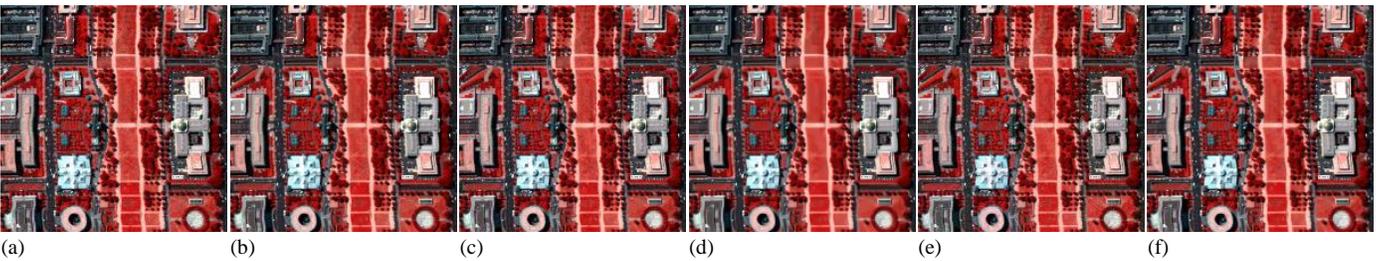

Fig. 10. Results for the Washington DC Mall image with $\sigma_n = Gau(200,30)$ in Case 3. (a) Pseudo-color noisy image with bands (57, 27, 17). (b) HSSNR. (c) LRTA. (d) BM4D. (e) LRMR. (f) The proposed method.

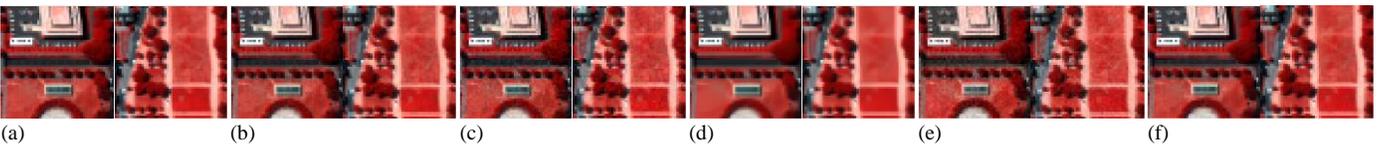

Fig. 11. Magnified results for the Washington DC Mall image in Case 3. (a) Noise-free image. (b) HSSNR. (c) LRTA. (d) BM4D. (e) LRMR. (f) The proposed method.

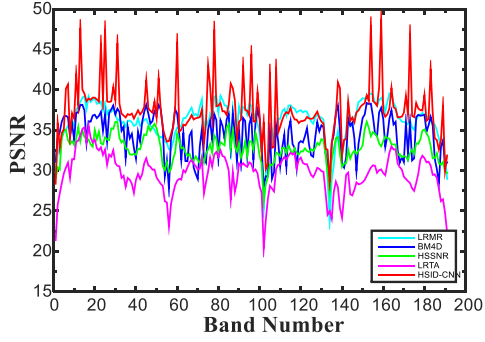 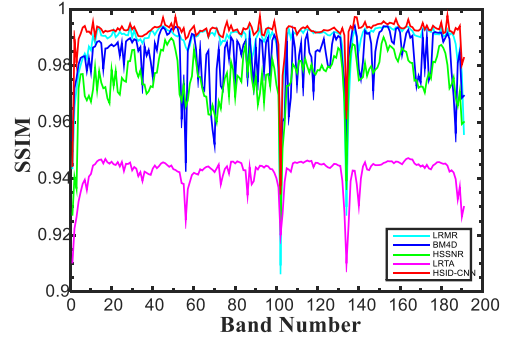

Fig. 12. PSNR and SSIM values of the different denoising methods in each band of the simulated experiment with noise level $\sigma_n = rand(25)$.

The spectral reflectivity is also crucial for HSI interpretation, such as classification, object detection, and unmixing [6], due to the physical properties of different ground objects. To validate the effectiveness after denoising in the spectral dimension with the different methods, Fig. 13 reveals the spectral curves of pixel (83, 175) in the restoration results of HSSNR, LRTA, BM4D, LRMR and the proposed method, respectively. The vertical axis named DN stands for the per-band value of the pixel in the same position, and the horizontal axis represents the band number. As displayed in Fig. 13, the proposed method outperforms HSSNR, LRTA, BM4D, and LRMR in terms of the performance in the spectral dimension and is closest to the ground truth.

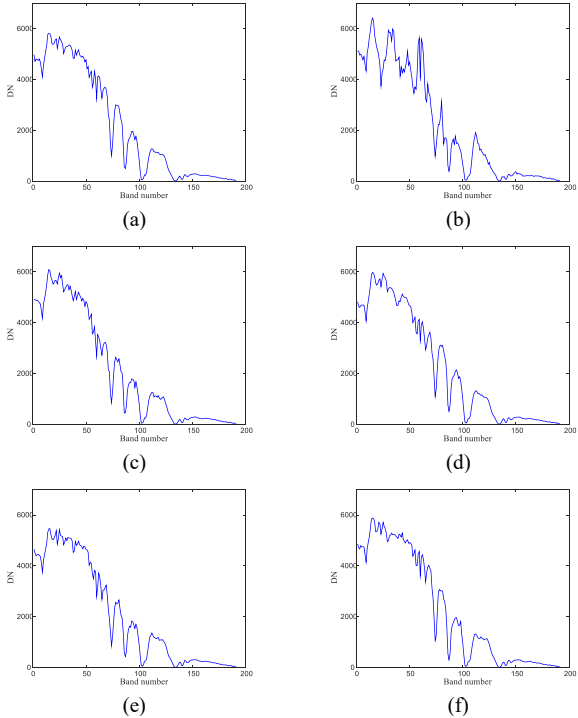

Fig. 13. Spectra of pixel (83, 175) in the restoration results. (a) Original. (b) HSSNR. (c) LRTA. (d) BM4D. (e) LRMR. (f) The proposed method.

In addition, to reveal the changes in the spectral reflectance after denoising, the spectral difference curves between the noise-free spectrum and the restoration results of the roof class at pixel (83, 175), grass class at pixel (105, 62), and road class at pixel (48, 120) are given in Fig. 14(a)–(c), respectively. In Fig. 14, the vertical axis of the figures represents the DN value difference between the restoration results and the noise-free HSI, and the horizontal axis represents the spectral band number. The difference curve of the proposed approach is smoother than the other algorithms for all three classes, with the residual value closer to zero, demonstrating that the presented method is more reliable in preserving the original spectral feature of the noisy HSI, as shown in Fig. 14.

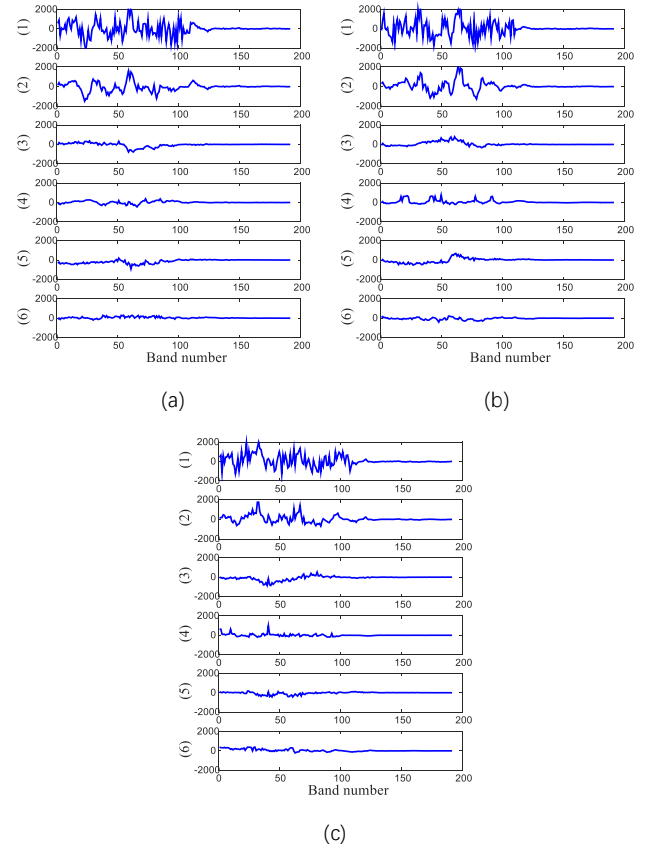

Fig. 14. Difference between the noise-free spectrum and the restoration results of (a) the roof class; (b) the grass class; (c) the road class. Curves (1)–(6) denote the results of the noisy image, HSSNR, BM4D, LRMR, and the proposed method, respectively.

## B. Real-Data Experiments

To further verify the effectiveness of the proposed method, two real-world HSI data sets were employed in our real-data experiments. The classification accuracy of the HSI before and after denoising is listed for comparison purposes with the different algorithms. Support vector machine (SVM) [51] was employed as the classifier under the same environment for all the restoration results. The overall accuracy (OA) and the kappa coefficient are given as evaluation indexes.

*1) AVIRIS Indian Pines Data Set:* The first few bands and several other bands of the Indian Pines HSI are seriously degraded by Gaussian noise and impulse noise [52]. Figs. 15 and 16 show the denoising results of contrast and the proposed method, which represent band number 2, and the pseudo-color result with combined bands (2, 3, 203), respectively.

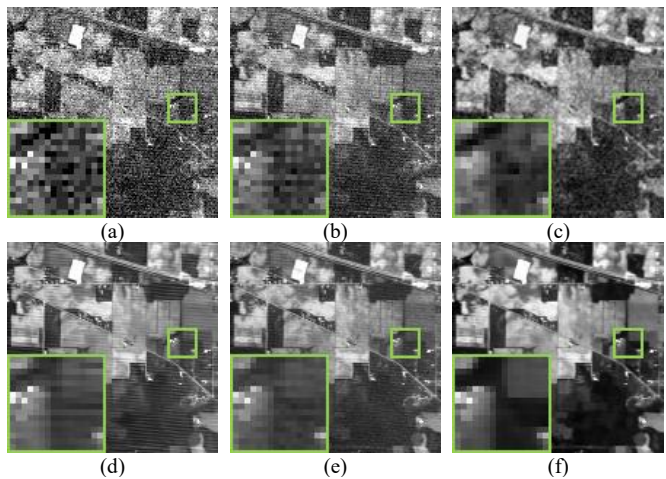

Fig. 15. Results for the Indian Pines image. (a) Real image band 2. (b) HSSNR. (c) LRTA. (d) BM4D. (e) LRMR. (f) Proposed.

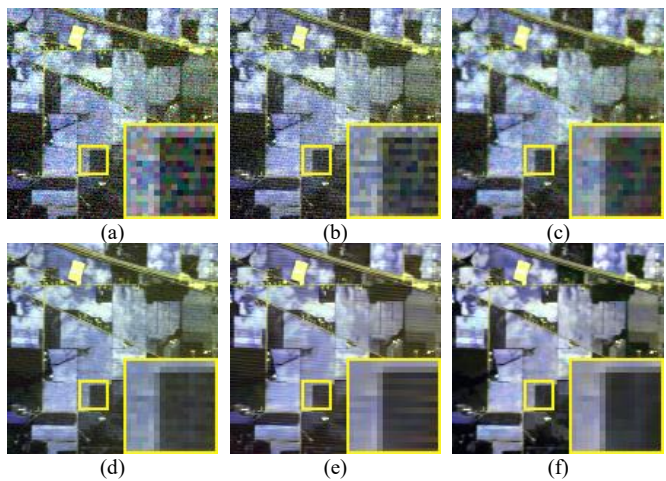

Fig. 16. Results for the Indian Pines image. (a) Pseudo-color noisy image with bands (2, 3, 203). (b) HSSNR. (c) LRTA. (d) LRMR. (e) BM4D. (f) Proposed.

In Figs. 15–16, it can be clearly seen that HSSNR can reduce some of the noise, but some dense noise and stripes still remain in the restored results. The LRTA method also shows the ability of noise suppression, but some detailed information is simultaneously smoothed and lost. BM4D does well in suppressing noise, but it appears to be virtually powerless against heavy striping. LRMR also behaves well in reducing both noise and heavy striping. However, the restored result of LRMR still shows obvious residual noise and stripes. HSID-CNN performs the best, effectively removing the noise and stripes, while simultaneously preserving the local details and structural information of the HSI.

In the supervised classification experiment with the SVM algorithm, 16 ground-truth classes were employed for testing the classification accuracy. The training sets included 10% of the test samples randomly generated from each class. The classification results with the Indian Pines image before and after denoising are revealed in Fig. 17. The OA and kappa coefficient are also given in Table III. Before denoising, as shown in Fig. 17(a), the classification results appear discontinuous, and the OA and kappa are only 75.96% and 0.7220, respectively. After denoising, as shown in Fig. 17(c)–(h), the OA and the kappa reveal different degrees of improvement. However, the classification results of HSSNR, LRTA, and LRMR still show an obvious fragmentary phenomenon, due to the noise removal of the original data being incomplete. BM4D, MH [53] and HSID-CNN suppress the fragmentary effect in most regions of the image, whereas HSID-CNN produces a better classification result, with the highest OA and kappa values of 85.65% and 0.8338, respectively.

TABLE III
CLASSIFICATION ACCURACY RESULTS FOR INDIAN PINES.

|       | Original | HSSNR  | LRTA   | BM4D   | MH     | LRMR   | Ours       |
|-------|----------|--------|--------|--------|--------|--------|------------|
| OA    | 75.96%   | 78.78% | 77.49% | 83.97% | 81.37% | 79.44% | **85.65%** |
| Kappa | 0.7220   | 0.7437 | 0.7387 | 0.8162 | 0.7895 | 0.7641 | **0.8338** |

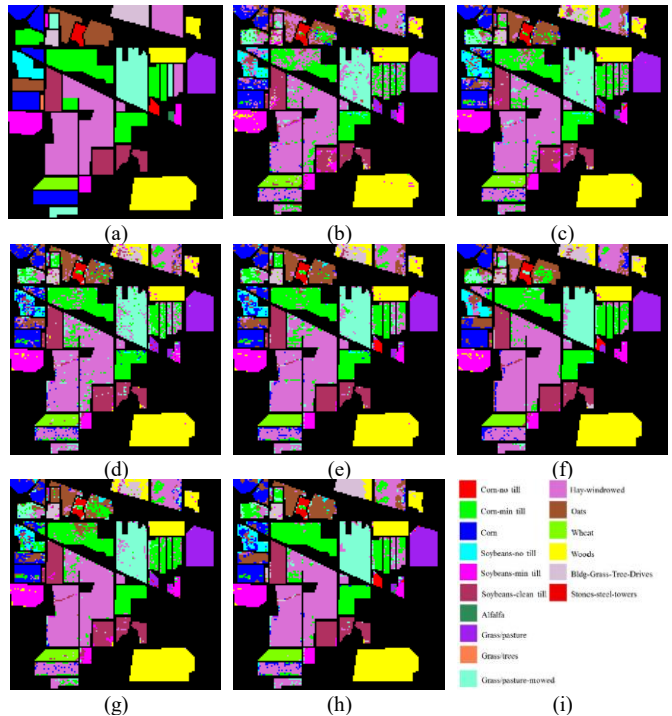

Fig. 17. Classification results for the Indian Pines image using SVM before and after denoising. (a) Ground truth. (b) Original. (c) HSSNR. (d) LRTA. (e) BM4D. (f) MH. (g) LRMR. (h) The proposed method. (i) 16 classes.

*2) ROSIS University of Pavia Data Set:* The noise is mainly concentrated in the first bands of the ROSIS University of Pavia HSI data. Figs. 18 and 19 show the denoising results of HSSNR, LRTA, BM4D, LRMR, and the proposed method, which represent the pseudo-color image with combined bands (2, 3, 97) and band number 2, respectively.

In Figs. 18 and 19, it can be clearly observed that LRTA cannot suppress the noise well. HSSNR and LRMR can reduce some of the noise, but some non-uniform noise still remains in the restored results. BM4D does well in suppressing noise, but it also introduces over-smoothing in some regions. HSID-CNN again performs the best, effectively removing the noise, while simultaneously preserving the local details and structural information, without obvious over-smoothing.

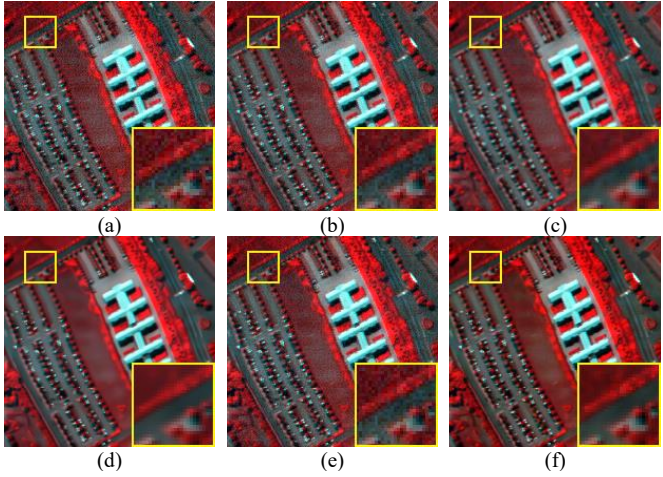

Fig. 18. Results for the Pavia University image. (a) Pseudo-color image with bands (2, 3, 97). (b) HSSNR. (c) LRTA. (d) BM4D. (e) LRMR. (f) Proposed.

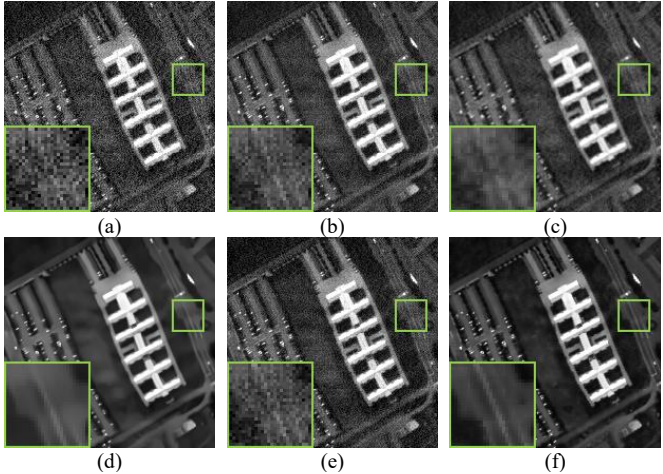

Fig. 19. Results for the Pavia University image. (a) Real image band 2. (b) HSSNR. (c) LRTA. (d) BM4D. (e) LRMR. (f) Proposed.

For the Pavia University HSI data, the noise is mainly focused in some of the first bands. Therefore, in order to better manifest the denoising effects of the different methods, the first 20 spectral bands were selected as the classification data. The classification accuracy results in Table IV also confirm the effectiveness of the proposed HSID-CNN, which acquires the highest OA and kappa coefficient values of 86.99% and 0.8319.

In Fig. 20, it can be clearly distinguished that the proposed method can reduce the fragmentary effect better than the HSSNR, LRTA, BM4D, MH and LRMR methods.

TABLE IV
CLASSIFICATION ACCURACY RESULTS FOR PAVIA UNIVERSITY.

|       | Original | HSSNR  | LRTA   | BM4D   | MH     | LRMR   | Ours       |
|-------|----------|--------|--------|--------|--------|--------|------------|
| OA    | 70.09%   | 71.66% | 72.56% | 78.88% | 84.87% | 83.95% | **86.99%** |
| Kappa | 0.6157   | 0.6373 | 0.6467 | 0.7302 | 0.8089 | 0.8148 | **0.8319** |

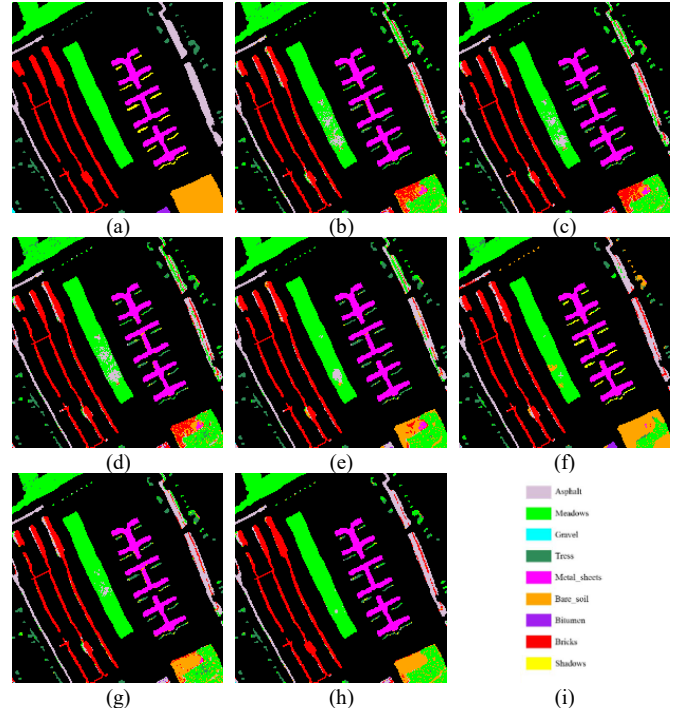

Fig. 20. Classification results for the Pavia University image using SVM before and after denoising. (a) Ground truth. (b) Original. (c) HSSNR. (d) LRTA. (e) BM4D. (f) MH. (g) LRMR. (h) The proposed method. (i) 9 classes.

*C. Further Discussion*

**1) Adjacent spectral band number $K$:** As described in Section III-B, the redundant spectral information in the HSI can be of great benefit to improve the precision of the restoration, since a spatial-spectral cube usually has a high correlation and similarity in the surface properties and textural features. Therefore, in the proposed framework, the current spatial band and its $K$ adjacent bands are simultaneously set as the inputs in the proposed network, employing a spatial-spectral strategy for HSI denoising. Hence, the adjacent spectral band number $K$ is a crucial parameter in the denoising procedure. In all of the simulated and real-data experiments, the number of adjacent spectral bands was set as $K = 24$. In fact, the choice of $K$ has a large effect on the restoration results of the proposed HSID-CNN method. To explore the influence of $K$ for HSID-CNN, Fig. 21 reveals the quantitative evaluation results (MPSNR) with different numbers of adjacent spectra $K$ (the horizontal axis represents a half value of $K$) in the simulated experiment. It can be clearly seen that the results of the proposed HSID-CNN method first quickly rise with the increase of $K$, and when $K = 24$, the results reach the highest

MPSNR value. The results then gradually descend with the increase of $K$. Clearly, the spatial-spectral strategy is significant for the proposed method.

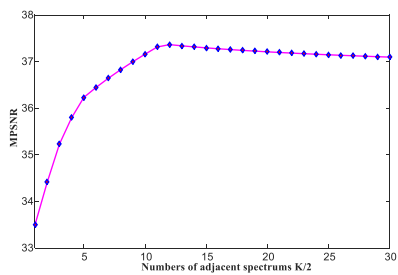

Fig. 21. Restoration results under different numbers of adjacent spectra $K/2$.

**2) Multi-scale feature extraction:** In the procedure for recovering the original information in HSI data, the feature expression may rely on contextual information in different scales, since ground objects usually have multiplicative sizes in different non-local regions in remote sensing imagery. Therefore, the proposed model introduces a multi-scale convolutional unit to extract more features for the multi-context information. To demonstrate the impact with/without multi-scale feature extraction, two comparison experiments were implemented with Indian Pines HSI data, as shown in Fig. 22. Some stripe noise is still residuary in the enlarged regions, where the model with multi-scale feature extraction performs better than the model without. This also certified that the proposed unit is beneficial for extracting multi-scale contextual information, which is critical and universal for diverse-resolutions HSIs denoising.

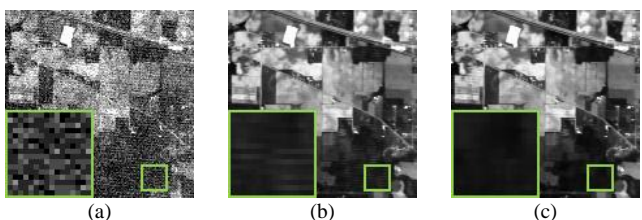

Fig. 22. With/without multi-scale feature extraction unit in Indian Pines HSI data. (a) Original. (b) Without multi-scale feature extraction unit. (c) With multi-scale feature extraction unit.

**3) Multi-level feature representation:** Due to the various levels of feature information in the different depth layers, as displayed in Fig. 4(b)–(e), it is worth merging these different feature representations for the final restoration. To efficiently transfer these comprehensive features between indirectly connected layers without attenuation, a multi-level feature representation unit is employed in the proposed HSID-CNN, as shown in Fig. 5. The unit integrality concatenates the multiple feature maps of the convolutional layers (layers 3, 5, 7, and 9) with different depths. To assess the impact on different levels of noise with/without multi-level feature representation, two comparison experiments were implemented with different noise levels, as shown in Fig. 23. With the increase of the noise intensity, the model with multi-level feature representation performs better than the model without. It can be clearly demonstrated that the proposed unit is beneficial for suppressing strong noise, through merging the different feature representations for the final restoration.

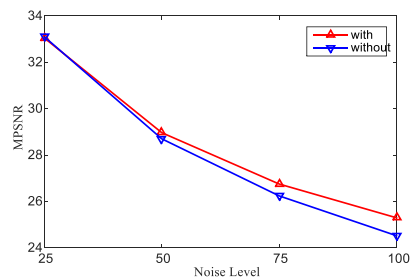

Fig. 23. HSI denoising results under different noise levels, with/without multi-level feature representation.

**4) Comparisons with DL-based Denoising Methods:** For further verifying the designed deep learning-based structure, we also compare with several CNN-based denoising method, such as DnCNN [42] and 3D extension of DnCNN with nine layers (3D-DnCNN). The contrasting evaluation indexes with noise level $\sigma_n = 25$ are listed in Table V. Fig. 24(b)-(d) show the denoising results with detailed parts of noisy image, DnCNN, 3D-DnCNN, and the proposed method, which represent the pseudo-color result, respectively. Due to the ignoring of spectral information, DnCNN only considers the spatial feature through band-by-band mode, which doesn't completely remove the spectral noise and damages spatial details for HSI data. Therefore, the authenticity of this single band-based denoising method is insufficient. This manifests that necessity of spatial-spectral strategy in HSIs processing. 3D-DnCNN employs the 3D convolutions, which takes the joint spatial-spectral information into consideration. Nevertheless, this model doesn't make allowances for the scale difference between ground objects in remote sensing data. In comparison, the proposed method both achieves the best evaluation indexes and visual effects, which demonstrate the superiorities of the combination with joint spatial-spectral strategy, multi-scale feature extraction and multi-level feature representation.

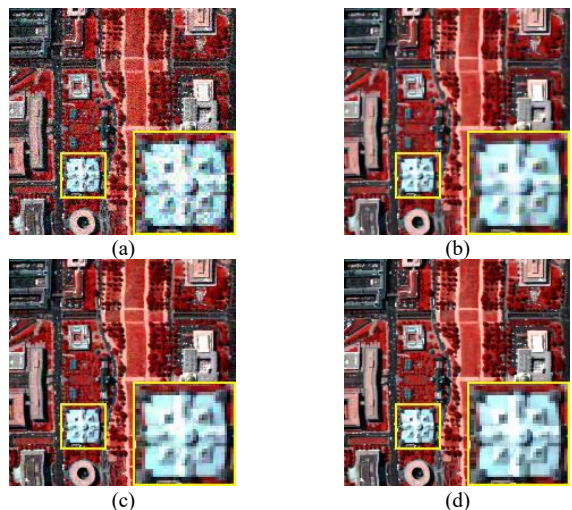

Fig. 24. Comparisons with deep learning-based HSI denoising methods. (a) Noisy. (b)DnCNN. (c) 3D-DnCNN. (d) Proposed.

TABLE V
COMPARISONS INDEXES WITH DL-BASED DENOISING METHODS.

| Index | DnCNN | 3D-DnCNN | HSID-CNN |
|---|---|---|---|
| MPSNR | 24.874 ± 0.0032 | 31.953 ± 0.0034 | **33.050 ± 0.0028** |
| MSSIM | 0.8805 ± 0.0003 | 0.9706 ± 0.0002 | **0.9813 ± 0.0001** |
| MSA | 9.7423 ± 0.0048 | 5.4274 ± 0.0036 | **4.2641 ± 0.0026** |

**5) Runtime Comparisons:** For evaluating the efficiency of denoising algorithms, we make statistics of average runtime in the simulated experiments under the same environment with MATLAB R2014b, as listed in Table VI. Distinctly, HSID-CNN exhibits the lowest run-time complexity than other HSI denoising algorithms with GPU mode, because of the high efficiency of end-to-end deep learning framework.

TABLE VI
AVERAGE RUNTIME COMPARISONS FOR HSI DENOISING METHODS IN THE SIMULATED EXPERIMENTS.

| Method | HSSNR | LRTA | BM4D | LRMR | Proposed |
|---|---|---|---|---|---|
| Time/s | 383.9 | 118.5 | 483.1 | 541.8 | **3.5** |

## V. CONCLUSION

In this paper, we have proposed a deep learning based HSI denoising method, by learning a non-linear end-to-end mapping between the noisy and clean HSIs with a deep combined spatial-spectral convolutional neural network (named HSID-CNN). Both the spatial information and adjacent correlated bands are simultaneously assigned to the proposed network, where multi-scale feature extraction is employed to capture both the multi-scale spatial feature and spectral feature. The simulated and real-data experiments indicated that the proposed HSID-CNN outperforms many of the mainstream methods in both evaluation indexes, visual effect, and classification accuracy of the denoising results.

In our future work, we will investigate more efficient learning structures to remove the mixed noise in HSI data, such as stripe noise, impulse noise, and dead lines [54]. Furthermore, another possible strategy which will be explored in our subsequent research will be to add *a priori* constraint or structure to the deep CNNs to reduce the spectral distortion and improve the texture details.

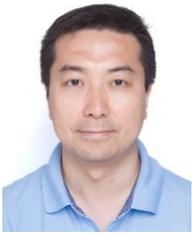

**Qiangqiang Yuan** (M'13) received the B.S. degree in surveying and mapping engineering and the Ph.D. degree in photogrammetry and remote sensing from Wuhan University, Wuhan, China, in 2006 and 2012, respectively.

In 2012, he joined the School of Geodesy and Geomatics, Wuhan University, where he is currently an Associate Professor. He published more than 50 research papers, including more than 30 peer-reviewed articles in international journals such as the IEEE TRANSACTIONS IMAGE PROCESSING and the IEEE TRANSACTIONS ON GEOSCIENCE AND REMOTE SENSING. His current research interests include image reconstruction, remote sensing image processing and application, and data fusion.

Dr. Yuan was the recipient of the Top-Ten Academic Star of Wuhan University in 2011. In 2014, he received the Hong Kong Scholar Award from the Society of Hong Kong Scholars and the China National Postdoctoral Council. He has frequently served as a Referee for more than 20 international journals for remote sensing and image processing.

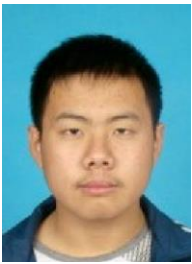

**Qiang Zhang** (S'17) received the B.S. degree in surveying and mapping engineering from Wuhan University, Wuhan, China, in 2017. He is currently pursuing the M.S. degree in School of Geodesy and Geomatics, Wuhan University, Wuhan, China.

His research interests include image quality improvement, data fusion, remote sensing image processing, deep learning and computer vision.

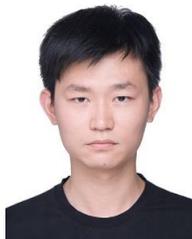

**Jie Li** (M'16) received the B.S. degree in sciences and techniques of remote sensing and the Ph.D. degree in photogrammetry and remote sensing from Wuhan University, Wuhan, China, in 2011 and 2016.

He is currently a Lecturer with the School of Geodesy and Geomatics, Wuhan University. His research interests include image quality improvement, image super-resolution reconstruction, data fusion, remote sensing image processing, sparse representation and deep learning.

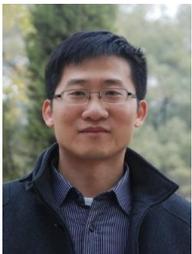

**Huanfeng Shen** (M'10–SM'13) received the B.S. degree in surveying and mapping engineering and the Ph.D. degree in photogrammetry and remote sensing from Wuhan University, Wuhan, China, in 2002 and 2007, respectively.

In 2007, he joined the School of Resource and Environmental Sciences, Wuhan University, where he is currently a Luojia Distinguished Professor. He has been supported by several talent programs, such as the Youth Talent Support Program of China in 2015, the China National Science Fund for Excellent Young Scholars in 2014, and the New Century Excellent Talents by the Ministry of Education of China in 2011. He has authored over 100 research papers. His research interests include image quality improvement, remote sensing mapping and application, data fusion and assimilation, and regional and global environmental changes. Dr. Shen is currently a member of the Editorial Board of the Journal of Applied Remote Sensing.

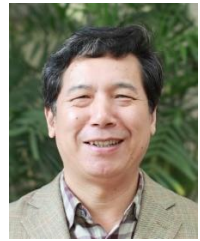

**Liangpei Zhang** (M'06–SM'08) received the B.S. degree in physics from Hunan Normal University, Changsha, China, in 1982, the M.S. degree in optics from the Xi'an Institute of Optics and Precision Mechanics, Chinese Academy of Sciences, Xi'an, China, in 1988, and the Ph.D. degree in photogrammetry and remote sensing from Wuhan University, Wuhan, China, in 1998.

He is currently the head of the remote sensing division, state key laboratory of information engineering in surveying, mapping, and remote sensing (LIESMARS), Wuhan University. He is also a "Chang-Jiang Scholar" chair professor appointed by the ministry of education of China. He is currently a principal scientist for the China state key basic research project (2011–2016) appointed by the ministry of national science and technology of China to lead the remote sensing program in China. He has more than 500 research papers and five books. He is the holder of 15 patents. His research interests include hyperspectral remote sensing, high-resolution remote sensing, image processing, and artificial intelligence.

Dr. Zhang is the founding chair of IEEE Geoscience and Remote Sensing Society (GRSS) Wuhan Chapter. He received the best reviewer awards from IEEE GRSS for his service to IEEE Journal of Selected Topics in Earth Observations and Applied Remote Sensing (JSTARS) in 2012 and IEEE Geoscience and Remote Sensing Letters (GRSL) in 2014. He was the General Chair for the 4th IEEE GRSS Workshop on Hyperspectral Image and Signal Processing: Evolution in Remote Sensing (WHISPERS) and the guest editor of JSTARS. His research teams won the top three prizes of the IEEE GRSS 2014 Data Fusion Contest, and his students have been selected as the winners or finalists of the IEEE International Geoscience and Remote Sensing Symposium (IGARSS) student paper contest in recent years.

Dr. Zhang is a Fellow of the Institution of Engineering and Technology (IET), executive member (board of governor) of the China national committee of international geosphere–biosphere programme, executive member of the China society of image and graphics, etc. He was a recipient of the 2010 best paper Boeing award and the 2013 best paper ERDAS award from the American society of photogrammetry and remote sensing (ASPRS). He regularly serves as a Co-chair of the series SPIE conferences on multispectral image processing and pattern recognition, conference on Asia remote sensing, and many other conferences. He edits several conference proceedings, issues, and geoinformatics symposiums. He also serves as an associate editor of the *International Journal of Ambient Computing and Intelligence, International Journal of Image and Graphics, International Journal of Digital Multimedia Broadcasting, Journal of Geo-spatial Information Science, and Journal of Remote Sensing,* and the guest editor of *Journal of applied remote sensing and Journal of sensors.* He is currently serving as an associate editor of the IEEE TRANSACTIONS ON GEOSCIENCE AND REMOTE SENSING.